\documentclass[sn-mathphys]{sn-jnl}

\usepackage{graphicx}
\usepackage{amsmath}
\usepackage{amsmath,amsfonts}

\usepackage{tabularray,booktabs}
\usepackage{multirow}
\usepackage{adjustbox}

\jyear{2021}%

\theoremstyle{thmstyleone}%
\theoremstyle{thmstyletwo}%

\theoremstyle{thmstylethree}%

\begin{document}

\title[LANDMARK]{LANDMARK: Language-guided Representation Enhancement Framework for Scene Graph Generation}


\author[1]{\fnm{Xiaoguang} \sur{Chang}}\email{xg\_chang@seu.edu.cn}

\author[2]{\fnm{Teng} \sur{Wang}}\email{wangteng@seu.edu.cn}
\author[2]{\fnm{Shaowei} \sur{Cai}}\email{shaoweicai@seu.edu.cn}
\author*[2,3]{\fnm{Changyin} \sur{Sun}}\email{cysun@seu.edu.cn}

\affil[1]{\orgdiv{School of Cyber Science and Engineering}, \orgname{Southeast University}, \orgaddress{ \city{Nanjing}, \country{China}}}

\affil*[2]{\orgdiv{School of Automation}, \orgname{Southeast University}, \orgaddress{ \city{Nanjing}, \country{China}}}

\affil*[3]{\orgdiv{School of Artificial Intelligence}, \orgname{Anhui University}, \orgaddress{ \city{Hefei}, \country{China}}}


\abstract{Scene graph generation (SGG) is a sophisticated task that suffers from both complex visual features  and dataset long-tail  problem. Recently, various unbiased strategies have been proposed by designing novel loss functions and data balancing strategies. Unfortunately, these unbiased methods  fail to  emphasize language priors in  feature refinement perspective.  Inspired by the fact that  predicates are highly correlated with  semantics  hidden in subject-object pair and global context, we propose LANDMARK (\textbf{LAN}guage-gui\textbf{D}ed representation enhanceMent frAmewo\textbf{RK})  that learns  predicate-relevant representations from  language-vision interactive patterns, global language context and pair-predicate correlation. Specifically, we first project object  labels to three distinctive semantic embeddings for different representation learning. Then, Language Attention Module (LAM) and Experience Estimation Module (EEM) process subject-object word embeddings to attention vector and predicate distribution, respectively. Language Context Module (LCM) encodes global context from each word embedding, which avoids isolated learning from local information.  Finally, modules outputs are used to update  visual representations and SGG model's prediction. All language representations are purely generated from object categories so that no extra knowledge is needed.  This framework is model-agnostic and  consistently improve  performance on existing SGG models. Besides, representation-level unbiased strategies endow LANDMARK the advantage of compatibility with other  methods. Code is available at \url{https://github.com/rafa-cxg/PySGG-cxg}.}




\maketitle

\section{Introduction}\label{intro}
Scene  graph generation (SGG) is a crucial task that benefits image captioning~\cite{ic1,ic3}, visual question answering~\cite{vqa3,vqa2,vqa1}, and video understanding~\cite{va1}. However, most generated scene graphs  faces challenge of trivial  predictions, thus far from been applied into practical applications.

Therefore,  recent  researches  have been working on unbiased methods that elevate  recall of hardly distinguishable predicates. Generally, unbiased methods can be divided into 3 types: data resampling (e.g., BLS \cite{BGNN}, GCL \cite{stracked}), predicate-aware loss design  (e.g., CogTree \cite{cogtree} and FGPL \cite{finegrain}) and logit manipulation (e.g., TDE \cite{unbias}, RTPB \cite{resistance}, FREQ \cite{motif}). However, a common drawback is that they rely on  explicitly modeling predicate correlations  or loss weights from dataset statistics \cite{finegrain,motif} or biased model predictions \cite{unbias,finegrain}, which means that they are sensitive to prerequisite changes. For instance, \cite{unbias} is not  effective when training on an unbiased model, hence, confining the SGG model performance. Compared with loss and statistic approaches,  language representation learning is  much robuster because it learns implicit patterns of predicates and avoids visual features redundancy, which is not been stressed by unbiased methods before. 

However, language representation learning has been adopted by some baseline models. For example,   \cite{attentiontranslation} takes word embedding to ground attention on visual features. \cite{stracked} utilizes Cross Attention (CA) mechanism for multi-modality learning.  \cite{sgnls} introduces transformer-based architecture to bridge
the gap between images and texts.  However, most of these approaches are not plug-and-play, and merely use single language representation  regardless of different semantic context. Therefore, failed to unleash power of language. 
\begin{figure}[]
	\centering
	{
		\includegraphics[width=0.48\textwidth]{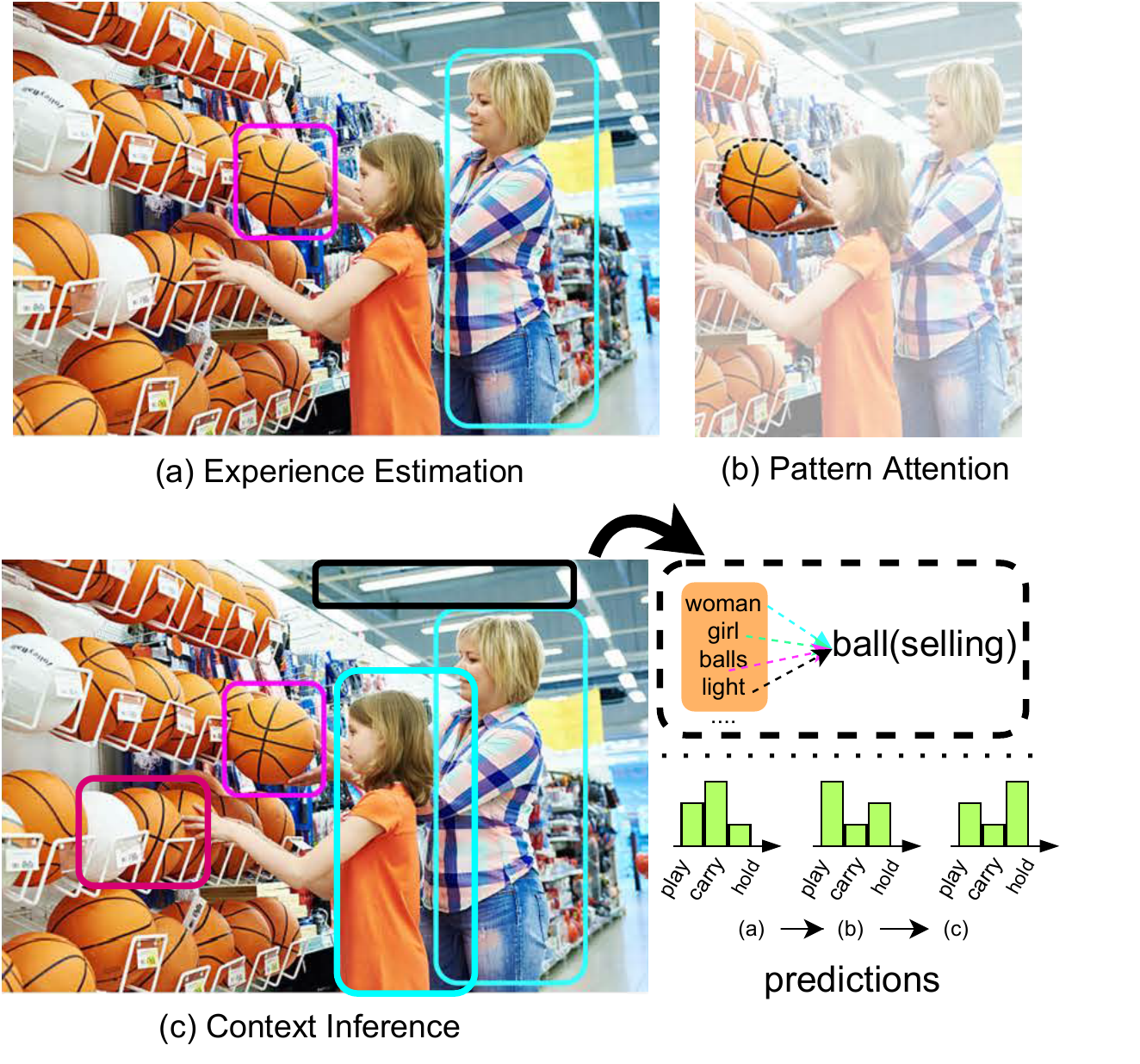}
	}

	\caption{An example of  multi-semantic language assistance for relation prediction.  Bottom-right corner shows  3 candidates'  possibility updating process.  a) \textbf{Experience Estimation}: human recalls a rough predicate distribution based on co-occurance possibility of predicates and object pair.   b) \textbf{Pattern Attention}: using internal relationship between  object pair and visual information for locating predicate-relevant visual pattern. c) \textbf{Context inference}:   using context to refine the  meaning of  subject and object, preventing prediction biases caused by isolated considering object pairs. The right-bottom shows prediction possibility changing through 3 processes} 

\label{fig:example}
\end{figure}

In fact, words have multiple meaning that  carries different priors  in terms of  different semantics, which  can guide scene graph generation. Here,  we give   a multi-semantic reasoning example in  Fig.~\ref{fig:example}. Given  this shopping picture, first, human constructs a predicate distribution  by correlation between  predicate and  the pair \textit{woman}-\textit{ball}  as well as  their relative position. This knowledge comes from experience and the process is  vision-independent.  Next,  still based on \textit{woman}-\textit{ball}, human can build correlation between  subject-object pair and visual pattern like ``woman's hand is closed to ball'', which  is a strong ``holding'' relevant pattern. In contrast, woman and girl's contour are irrelevant in terms of this pair.  Finally, according to surrounding objects (e.g., balls, girls, lights), human can  infer   the   \textit{selling} context  to avoid predicting  \textit{play}, because it is not suit in this scene. 

Motivated by these observations, we heuristically design 3 plug-and-play language modules that exploit different language prior behind object categories.  Three modules take detected object classes as input and generate  semantic embeddings into different semantic spaces, which are used for extracting priors from language-visual pattern correlations, language context and pair-predicate correlation, respectively. Concretely, 1) \textbf{Language Attention Module} projects subject-object word embeddings to unified semantic matrix, then,  channel attention  is used to extract attention vector  for relation visual feature map, which can learn relevance between object pair and specific predicate-relevant visual patterns  2) \textbf{Language Context  Module} employs transformer-based encoder to encode the global language context into entity's semantic embedding from a sequence of entity labels. Comparing with  pretrained word embedding, this module can generate semantic representation that fits  context. These two modules are used for initializing entity and relation visual-based  representations, respectively. 3) \textbf{Experience Estimation Module} are supervised by  marginal probability of subject-predicate and object-predicate  to learn the class and spatial aware predicate distribution  as likelihood offset. It is worth mentioning that language processing is disentangled with visual feature at very beginning, so this framework is applicable for most SGG baseline models. 

To best of our knowledge, we are the first to utilize multi semantic language representation within object label to achieve unbiased Scene graph generation. The main contributions could be summarized as follows:
\begin{enumerate}
	\item We  propose  LANDMARK that introduces language representation learning into unbiased scene graph generation, which stresses the under-explored multiple semantics utilization in object label.
  \item  We devise three modules that divide object labels into distinctive semantic spaces, then extract  priors of  language-vision interactive patterns and  semantic context as well as pair-predicate correlation, respectively. 
\item  Experiments  on SGG branchmark show constant improvements on  baseline models and compatibility with other unbiased methods,  which  indicate the effectiveness  of multi-semantic language representations that induced from   object labels. 

\end{enumerate}
\section{Related works}
\label{sec:1}
\textbf{Scene Graph Generation:} There are mainly two
mainstream methods for scene graph generation: Based on context modeling
or graph convolutional network (GCN) \cite{gcn}. The first approach is focused on modeling global information and widely adapted message passing between all entities features \cite{Gps-net,mem,linknet,factorizable,structure}. A number of them \cite{motif,phrases,stracked,resistance,sgnls} harnessed sequential  models e.g., LSTM \cite{lstm}, GRU \cite{gru}
and Transformer \cite{attention}. Chen {et al.} \cite{resistance} use two stacks of Transformer to encode global
information. Zellers {et al.} \cite{motif} uses LSTM to encode global context that informs relation prediction. However, merely modeling global context is not sufficient for Scene graph tasks.
Another approach \cite{BGNN,energy,grcnn,drnet} propagate massage between node and edge features,
and focusing more on regional pair-wise information. \cite{BGNN} applied a multi-stage graph
message propagation between proposal entities and relationship representations.
In \cite{grcnn}, Yang \textit{et al.} pruned graph connections to sparse one,
then attentional graph convolution network is applied for modulating
information flow. \cite{energy} utilized GCN for updating state representations
as energy value.  Chen \textit{et al.}\cite{kern} constructed a graph between
proposal and all relationship representations and aggregated messages by GRU.
Yet, this approach  suffers from  insufficient global context encoding. Our method considers both pair-wise and global contexts for representation refinement.

\noindent\textbf{Language prior learning:} entity labels and relation distributions are most used prior knowledge in scene graph generation. On unbiased method side, language knowledge can be embodied in loss weight or prediction bias. On For example, FREQ \cite{motif} directly counts ``subject-predicate-object'' co-occurrence in dataset as prediction bias.  Chen et al. \cite{resistance} devises different
types of specific bias by using different ways to model object
relationship from dataset. CogTree \cite{cogtree} proposed  a loss based on  automatically built
 cognitive structure of the relationships from the biased SGG predictions. Through mentioned methods get remarkable boost on specific baseline models in terms of metric mRecall@k \cite{vctree}, they are instable when prerequisite changes.

On the baseline model side, an increasing number of works strived for
multi-modelity representation learning. \cite{lp} designed a
language module, which projected representation to the same space of vision module
for minimizing the distance of similar semantic features.
\cite{vctree} introduced a confidence estimation module to alleviate the error
propagation by incorporating confidence estimation in node feature
updating. 
\cite{bridge} incorporated external commonsense knowledge by unifying
the formulation of scene graph and commonsense graph.
\cite{commonsense} structured visual commonsense and proposed a cascaded fusion
architecture for fusion. However, existing methods treat language modality as a single representation, which lost a lot of information.  
\begin{figure}[t!]
	\includegraphics[width=0.98\textwidth]{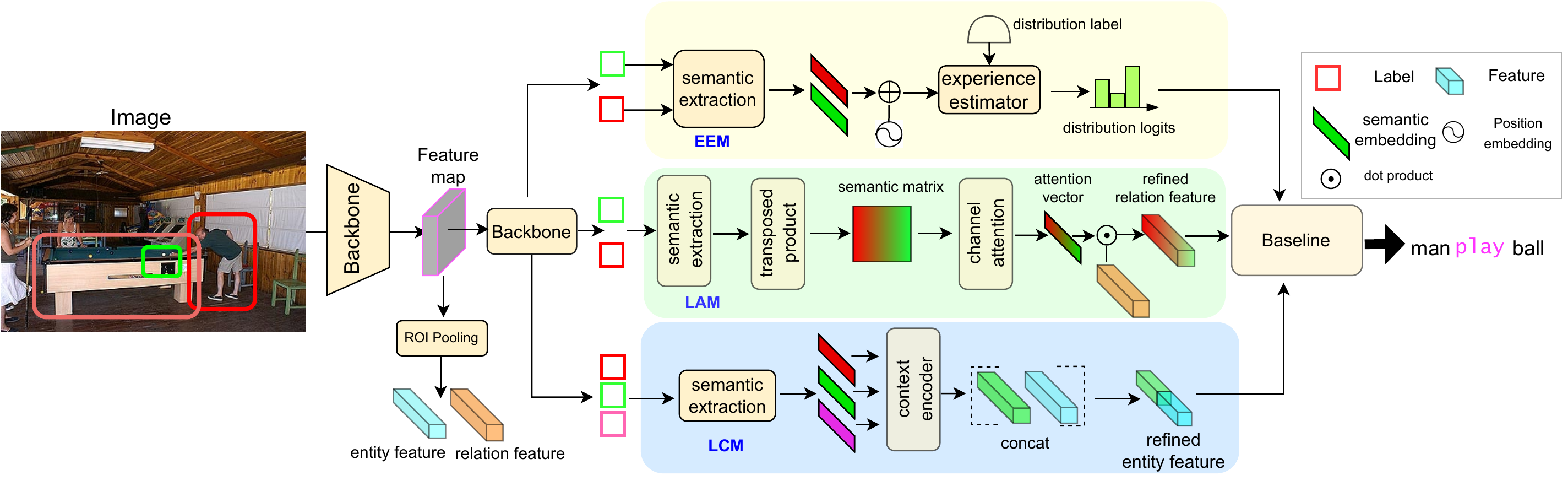}\
	\caption{LANDMARK architecture. The image first goes through a generic object detector to get predicted object's label and ROI feature. Then labels are served as  three modules input (i.e., EEM, LAM, LCM) and calculated distinctive semantics. Finally, LAM and LCM outputs are used to refine relation and entity representations, respectively. EEM's output is served as prediction offset.}
	\label{net}
\end{figure}
\section{Methodology}
\label{sec:2}
\textbf{Problem formulation:}  Given an image $I$, scene graph generation aims to predict entity class set $\mathcal{C}_{e}$, coordination set $\mathcal{B}_{e}$ and relation set $\mathcal{C}_{r}$. Generally, existing SGG models receive visual entity and predicate (or node and edge) representations from backbone. Then  a graph $\mathcal{G} = \{\mathcal{C}_{e},\mathcal{B}_{e}, \mathcal{C}_{r}\}$ can be formulated. 
\begin{equation}
	\mathcal{G}=P(\mathcal{C}_{r},\mathcal{B}_{e},\mathcal{C}_{e}\vert I)=SGG({N},{E})  
\end{equation}
Where ${N}= \{ e_{i}\}_{i=1}^{n} $ and ${E}=\{ e_{ij}\}$ are the set of entity and predicate representations. In this paper, we aimed to update $N$ and $E$ by incorporating semantic priors. \\

\noindent \textbf{Framework overview:} LANDMARK  consists of three  semantic learning modules, i.e.,   \textit{Language Attention Module} (LAM), \textit{ Language Context Module} (LCM) and \textit{Experience Estimation Module} (EEM). The  framework architecture is shown in Fig.~\ref{net}. First, we obtain  ${N}$, ${E}$ from  ROI Pooling, $\mathcal{C}_{e}$,  $\mathcal{B}_{e}$ from classifier head.
Then, semantic extraction operation convert labels  $\{ c_{i}\} $ to semantic embeddings for each module. For LAM, semantic embeddings of  subject $c_{i}$ and object $c_{j}$ are transposed and multiplied  to semantic matrix, then  channel attention transfers the matrix to attention vector and update relation representation. 
{LCM} encodes semantic embeddings of all entity labels  $\mathcal{C}_{e}$ presented in the image, generating  context-aware semantic entity feature and concatenating it with visual entity representation. The updated  entity and relation  representation   are passing through the baseline model.
For EEM, the distribution label is generated to supervise experience estimator, which  combine subject-object semantic embedding with position embedding to yield distribution logits. Finally, generated logits are used to update the final predicate likelihood.

\subsection{Semantic extraction}
\noindent Semantic extraction is used to transfer labels to the corresponding semantic space, which is applied on three modules independently. Specifically, semantic extractor consists of three operations:

\begin{align}
     f_{se}^{sub}(c_{i})&=w^{T}_{s}c_{i} \nonumber \\
    f_{se}^{obj}(c_{j})&=w^{T}_{o}c_{j} \nonumber \\
    f_{se}^{ent}(c_{e})&=w^{T}_{e}c_{e}
    \label{eq2}
  \end{align}

The first and second operations  are used in LAM and EEM for subject and object projection respectively. Considering that the same word as subject or object may have contrastive meaning (e.g., \textit{eating} could be a possible predicate if ``man'' is subject, which is impossible  when ``man'' is object),  we use different weights $w_{s}$ and $w_{o}$ to project  subject and object to semantic embedding.  Last operation is used in LCM, since all labels are treated  as  objects, we use unified $w_{e}$ as semantic embedding weight. 

In fact, $f_{se}$ could be any projection function as long as  the input is object labels. Here, we only use a naive 1-layer linear function to prove  the extraction's effectiveness.

\subsection{Language Attention   Module}
\label{sec:lmm}
\noindent This module aims to learn the   prior between object pair  and visual predicate-relevant patterns within visual relation representation $e_{ij}$. Original feature extraction network (e.g., Reset \cite{resnet}) keep both spatial and semantic information. Specially, different  channels focus on different visual patterns. However, relation visual feature inevitably mix up with  huge amount of  irrelevant background information, so there is a  need for channel selection.  Heuristically, given a specific subject-object pair (e.g., \textit{boy}-\textit{basketball} or \textit{boy}-\textit{street}), visual feature should have different activation. Therefore, we design a label-aware channel attention mechanism. Specifically, given subject i and object j, we first generate a  semantic matrix $x_{ij}$ as a unique representation of word-vision correlation. 
\begin{equation}
	\label{in}x_{ij}=f_{se}^{sub}(c_{i}) \otimes f_{se}^{obj}(c_{j})^T  
\end{equation}
	
Where $\otimes$ refers to matrix multiplication.  We achieve channel attention by a series of 2D convolutions with the spatial pooling  on  $x_{ij}$ to get attention vector  $e_{ij}^{c}$:
\begin{equation}
	e_{ij}^{c} =\sigma(G_{\operatorname{pooling}}(G^{n_{c}}_{\operatorname{conv}}...\sigma
	(G^{1}_{\operatorname{conv}}(x_{ij}))))\in\mathcal{R}^{C,1}
	\label{eq:lm}
\end{equation}
Where $C$ is as  the channel number of visual relation feature $e_{ij}$, $n_{c}$ is the number of 2D convolution layers, $G_{\operatorname{pooling}}$ is pooling operation, $\sigma$ is activation function. Finally, the channel weights $e_{ij}^{c}$ will be used for updating $e_{ij}$, so that irrelevant channel for relation discrimination will be suppressed.
\begin{equation}
	\label{edge} 
	\hat{e}_{ij}= e_{ij} \times  e_{ij}^{c} 
\end{equation}
Where $\hat{e}_{ij}$ is refined relation representation, $\times$ is dot product operator.    

\begin{figure}[]
	\centering
	{
		\includegraphics[height=6cm]{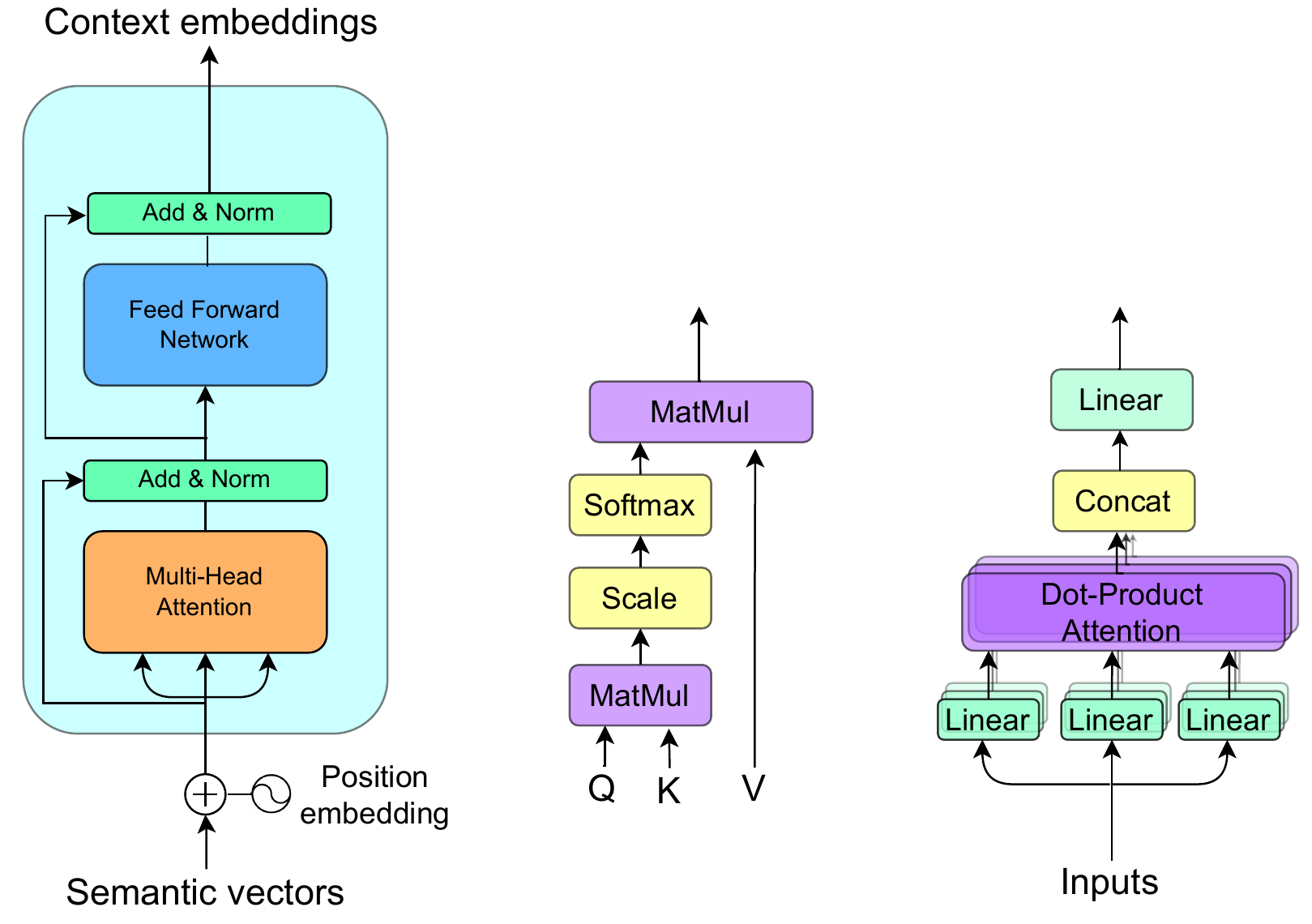}
	}

	\caption{ Context  Encoder block structure (left).  Dot-Product Attention (middle).  Multi-Head Self Attention (right).} 

\label{fig:encoder}
\end{figure}
\subsection{Language Context  Module}

\label{sec:sem}
\noindent Compared with visual information,  single word  is semantically isolated from other components in a sentence. Though we devise LAM and EEM for semantic extraction, the utilized  pairwise labels  are confined to local semantics, which is insufficient for comprehensive semantic inference. Hence, LCM is aiming at  addressing the global context deficiency problem. This module includes semantic extractor and  context encoder.  The context encoder consists of multilayer transformer encoder with Multi-Head Self-Attention (MHSA)~\cite{attention} and  Feed Forward Network (FFN) \cite{attention}. The structure is illustrated in Fig.~\ref{fig:encoder}. Concretely, given an image, supposed there are $n$ entities, the input sequence $X$ could be described as follows:
\begin{equation}
X = \left\{s_{0},s_{1},...,s_{n}\right\}
\end{equation}
 where
\begin{equation}
	s_{i}= \gamma (f_{se}^{e}c_{i} + p_{i}) \in \mathbb{R}^{d}
	\label{att_input}
\end{equation}
Here, $c_i$ and $p_{i}=\phi[x_{i},y_{i},w_{i},h_{i}]$ refer to class labels and entity's position embedding of the $i$-th entity. $x_{i}, y_{i}, w_{i}, h_{i}$ are  center coordinates, width, height of the object i.   $\gamma$ denotes the learnable linear transformation.  Where $d$ is the dimension of each element in the sequence.
 We first reiterate standard Scaled Dot-Product Attention~\cite{attention}. 
\begin{equation}
  \operatorname{Attention}(Q, K, V)=\operatorname{softmax}\left(\frac{Q K^T}{\sqrt{d_k}}\right) V
  \end{equation}
Then, Multi-Head Self Attention is formulated as:
\begin{align}
\operatorname{MHSA}(X) &=\text { Concat }\left(\operatorname{head}_1, \ldots, \operatorname{head}_{\mathrm{h}}\right) W^O \nonumber \\
\text{head}   &=\text { Attention }\left(X W_i^Q, X W_i^K, X W_i^V\right)
\end{align}

 Where $W_i^Q , W_i^K, W_i^V$ and $W^O $ are  parameter matrices. The $b$-th layer output $X_{b}$ can be denoted as:
\begin{align}
X_{b}' = \operatorname{MHSA}(\text{LN}(X_{b-1})) &+ X_{b-1},
\\
X_{b} = \operatorname{FFN}(\text{LN}(X_b')) &+ X_{b}', 
\end{align}
Where $X_{b-1}$ is the $(b-1)$th layer,  and we set $X_0 = X$. LN is layer normalization. Differing from previous works~\cite{seq2seq}\cite{resistance}, LCM takes a sequence of entity labels  as inputs, so module can learn semantic entity representation on top of high level information.

\subsection{Experience Estimation  Module:}
\noindent This module is  deigned to supervised learn the relationship distribution prior from  subject-object pair,  as a compensation of cross entropy loss. EEM consists of semantic extractor, experience estimator and distribution label generation for supervision. Since entity class and position  both have  influence on judging relation, This module utilizes both  classes and position information to learn precise distribution. First, we embed entity label i, j and position embedding $p_{ij}$ to high dimension representation space, then, experience estimator  predict the  predicate distribution $d_{ij}$ between subject i and object j.

\begin{align}
	\centering
	p_{ij} & =\phi_{p}[x_{i},y_{i},w_{i},h_{i},x_{j},y_{j},w_{j},h_{j}]\label{pos
	encoding}                                                                    \\
	{d}_{ij} & =\varphi\left[\phi_{s}(f_{se}^{sub}(c_{i}))
	\phi_{o}(f_{se}^{obj}(c_{j})),p_{ij}\right]
\label{est}
\end{align}

Where $[\cdot , \cdot]$ refers to concatenation operation,  $\phi_{p},\phi_{s},\phi_{o},\varphi$ are fully connected layers with RELU as activation function. Finally, we merge relationship distribution $d_{ij}$ with the prediction from the baseline, which could be described as follows:
\begin{equation}
\hat{d}_{i,j} = SGG(\hat{n}_{ij},\hat{e}_{ij}
)+d_{ij}
\end{equation}
Where $\hat{d}_{i,j}$ is the updated prediction likelihood. $\hat{n}_{ij}$  and $\hat{e}_{ij}$ are enhanced entity and relation feature  obtained by LAM and  CAM (Section~\ref{sec:lmm}, \ref{sec:sem}).

\begin{figure}[tbp!]
        \centering
        {
            \includegraphics[width=0.5\textwidth]{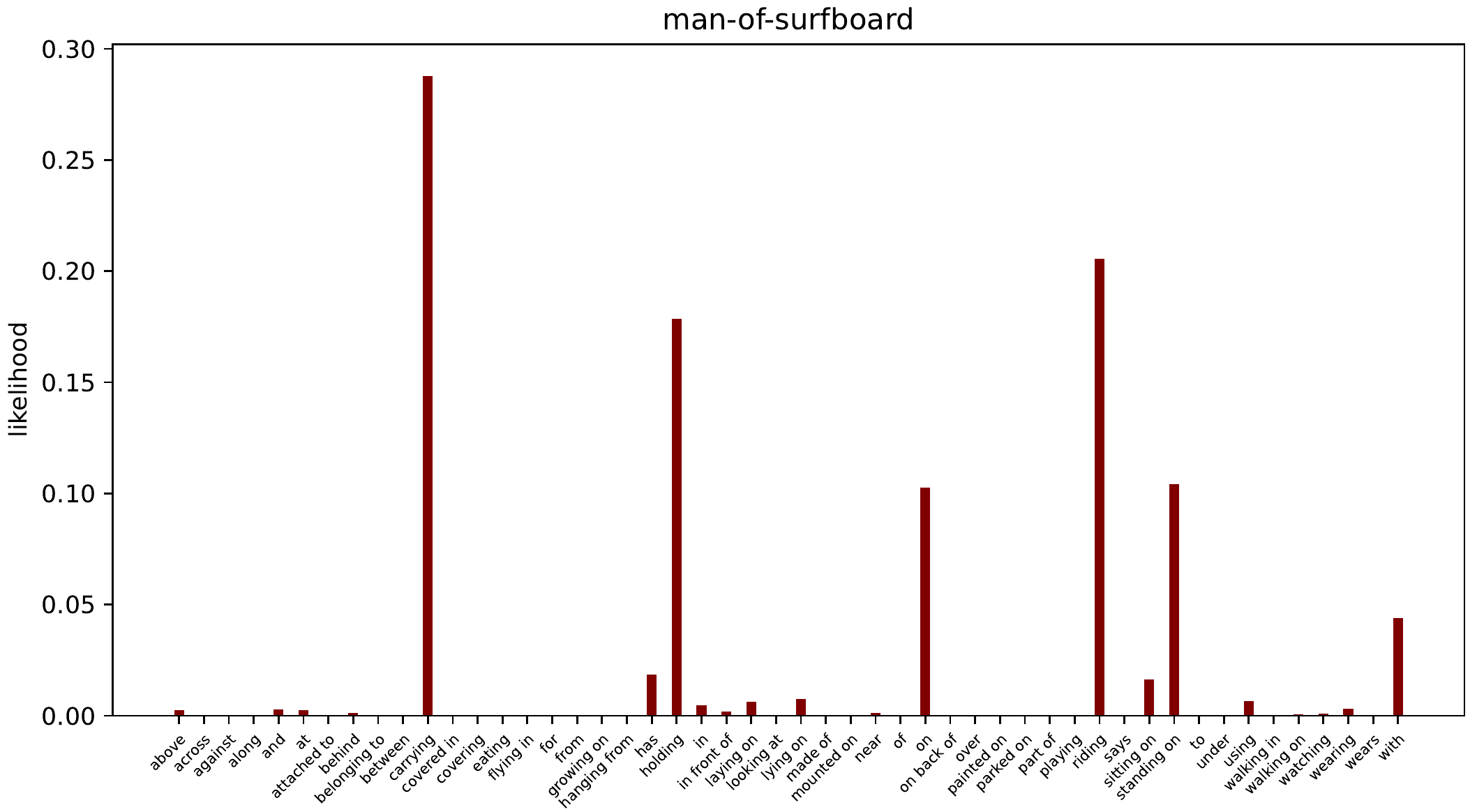}
        }
        \caption{Joint possibility for subject \textit{man} and object \textit{surfboard}. The groundtruth triplet $\langle \text{man,of,surfboard}\rangle $ is sampled from zero-shot split (where triplets are only in evaluation set but not occurred in training set).  } 
        \label{fig:distribution}
    \end{figure}

\noindent \textbf{Distribution label generation:} Dataset annotations  are inherently reflect  human commonsense, so we manage to generate accurate distribution labels from dataset.    For subject i, object j, we obtain  $m^{sub}_{i}$ $ m^{obj}_{j}$  as  ``subject-predicate'' and ``predicate-object''  marginal distributions. Since $m^{sub}_{i}, m^{obj}_{j}$ are independent distribution, we calculate joint possibility  $p^{joint}_{ij}$ as followed:  
\begin{equation}
	p^{joint}_{ij}  =m^{sub}_{i} \odot m^{obj   }_{j}    
\end{equation}
Where  $\odot$ denotes the element-wise product. Though EEM is alike FREQ \cite{motif} that  generating predicate distribution from statistics, FREQ  directly counts  triplets  $\langle subject, predicate, object \rangle$  occurrence. However,  some triplet samples are scarce in training samples, so it is hard to establish an informative distribution prior.  In contrast, EEM uses joint possibility as labels to infer the predicate distribution. Fig.~\ref{fig:distribution} is a typical example of generated joint possibility  $p^{joint}_{ij}$ of a triplet that not occurred in training set. 	In this circumstance, FREQ could not work due to zero sample number, whereas, we can see that top 5 highest likelihoods  are reasonable and include many possible scenarios. In contrast, the predicate ``of''  is ambiguous. Better predicate could be one of top likelihoods from joint possibility.

Considering that introducing position information makes accurate predicate possible,  we design a fusion function for mitigating joint possibility and true predicate label. The distribution label $l_{ij}$ can be denoted as:
\begin{equation}
    l_{ij}  = \mu \times p^{joint}_{ij} + (1-\mu)\times\text{onehot}[r_{ij}]
    \label{eq:miu}
    \end{equation}
    Where $\mu$  is a factor regulating proportion of marginal frequency.   $r_{ij}$ refers to real predicate label  between subject $i$ and object $j$. 
    
    \noindent \textbf{Objective function:} we choose MSE loss 
    between distribution label $l_{i,j}$ and predicted distribution ${d}_{i,j}$, which is
    denoted as:
    \begin{equation}
        \text{MSE} = \frac {1}{ K }\sum _{i=1}^{K}
        ({d}_{i},l_{i})^{2}
        \end{equation}
        Where K is the number of relation categories in dataset.
   
    \subsection{Baseline Model}
    \noindent Any off-the-shelf two-stage scene graph
        generation model can be used as baseline. It could be either a sequential model
        or a graph neural network, as long as it exerts entity and relation features for
        prediction.
        \section{Experiments}
\label{experiments}
In this section, we first introduce the experiment settings in our experiments. Then, we test our framework's effectiveness on several SGG models, and conduct experiment  to analyze the  compatibility between our method and the state-of-art unbiased  strategies.  Finally,  detailed analyzations and quantitative evaluations are presented to further verify  LANDMARK in different perspectives.
\subsection{Experiment settings}

\noindent \textbf{Datasets}  We  employ the widely adopted  {Visual Genome~\cite{visualgenome} filtered subdataset VG150~\cite{imp}}  to train and evaluate our framework. The VG150 dataset
contains the most frequent 150 object categories and 50 predicate categories in VG. It consists of more than 108k images, with 70\% images held out for training and 30\% for testing. Among the training set, 5000 images are used for evaluation.

  \noindent\textbf{Tasks.} We consider three conventional sub-tasks of scene graph generation to evaluate our framework. 1) \textit{Predicate Classification (PredCls)} predicts relationships between each object pairs given their ground-truth bounding boxes and classes. 2) \textit{Scene Graph Classification (SGCls)} predicts the object classes and their  relationships given the ground-truth bounding boxes of objects. 3) \textit{Scene Graph Detection (SGDet)} needs to detect object classes, bounding boxes and predict their relationships.\\

  \noindent\textbf{Evaluation Metrics}   we report widely accepted metrics Mean Recall@k  \cite{kern} and Recall@k \cite{vctree} for evaluating models performance.   Recall@k computes the fraction of times the correct relationship is predicted
  in the top k  relationship predictions. However, the R@K fails to reflect the performance on tail relationship categories, thus,  Recall on each class (i.e., mean Recall) is used to evaluate unbiased performance. Besides, In order
  to evaluate EEM, we need a metrics to measure distribution accuracy, so that TOP-N Recall@K metric are purposed by allowing top N scored predicate in one prediction as candidates, then, $N\times K$ number of candidates are used for calculating Recall, i.e.
  \begin{equation*}
    \text{TOP-N Recall@K} = \frac{{correct}(\text{{\{ candidates\}}}_{N\times K})}{N_{gt}}
    \end{equation*}
  When N=1, TOP-N Recall@K is equal to Recall@K. The difference between TOP Recall and Recall with no graph constraint \cite{pixels} is the former one forcing top $N\times K$ scored predicates to be preserved, avoiding candidates been excluded by higher scored in other predictions.

  \noindent\textbf{Implementation Details}  We use pretrained Faster R-CNN \cite{rcnn}  with backbone  ResNeXt-101-FPN \cite{maskrcnn} and ROIAlign \cite{maskrcnn} as object detector. We froze its weights during training. For our framework, we use pretrained GloVe \cite{glove} weight as initial $w_{s},w_{o},w_{e}$ in semantic extractor. For EEM, we use 3 layers MLP $\Phi_{s},\Phi_{o},\Phi_{p}$
  with 1024 neurons, $\varphi$ is a 2 layers MLP with hidden dimension 4096. $\mu$ referred in Eq.~\ref{eq:miu} are set to 0.3 for unbiased methods, 0.7 for baseline models. For Eq.~\ref{eq:lm}, we choose two $3\times3$ convolution layers to generate a 256-channel attention vector. For LCM, We choose context encoder with 4 layers and 8 heads, entity dimension $d=512$. For training, approximately 10000 iterations is enough for each baseline. The basic learning rate is 0.01 and batch size is 16. We choose the SGD optimizer for optimization.

    \subsection{Comparison with baseline methods} 
  \noindent Table \ref{tab:eval} shows mRecall \& Recall of 5 baseline  models with or without our framework LANDMARK. Baseline  models include GCN based models: G-RCNN~\cite{grcnn}, BGNN~\cite{BGNN} and context modeling networks: IMP~\cite{imp}, Transformer~\cite{attention},  Motifs~\cite{motif}.   It is worth mentioning that we do not deploy any unbiased  strategies on these models. We observe that incorporating our proposed LANDMARK leads to a consistent mRecall improvement  in all three tasks for all  baseline models, which demonstrates the robustness of our approach. For mR@100, our model average improvements are  3.66\%, 2.64\%, 1.37\% on three tasks. The improvements  might be attributed to the fact that  multi-semantic language  representation indeed facilitates visual representation.  It is not surprised that average improvement consecutively shrinks in three tasks, due to inaccurate class and position predicted by pretrained object detector.   Besides, Recall shows drops in different extent, which is a common characteristic of an unbiased method.

  \textcolor{black}{We also measure the total parameter amounts (M)  and computation overhead (GFLOPs) of baseline and LANDMARK in   Table \ref{tab:eval}. Generally, there is ~20M (relative:  5.3\%) parameters and ~2.5GFLOPs (relative: 1.0\%) computation increased. In comparison with BGNN+TDE (365.7M \& 213.2GFLOPs), both model size and computation overhead are smaller.  That is attributed to lightweight module design and adoption of low-dimensional inputs (i.e., language rather than image inputs.)}

  \begin{table*}[t]
	\centering
	\caption{The SGG performances of mRecall \& Recall  on VG dataset in \%. * denotes the default unbiased strategy in paper is not applied. All  baselines are reimplemented on our codebase.}
	\label{tab:eval}
	\resizebox{\linewidth}{!}{%
	\begin{tabular}{ccccccccc} 
	\hline
  \multirow{2}{*}{Model} & \multirow{2}{*}{Method} & \multicolumn{2}{c}{PredCls} & \multicolumn{2}{c}{SGCls} & \multicolumn{2}{c}{SGDet}  &  \multirow{2}{*}{\textcolor{black}{Params(M)/GFLOPs}} \\
	 &  & mR@50/100 & R@50/100 & mR@50/100 & R@50/100 & mR@50/100 & R@50/100 \\ 
	\cline{1-9}
	\multirow{2}{*}{IMP \cite{imp}} & baseline & 16.88/18.02 & 66.8/68.25 & 7.57/8.08 & 38.9/40.17 & 6.00/7.30 & 27.24/34.24 &336.3/206.4\\
	 & LANDMARK & \textbf{19.54/21.06} & 64.89/66.61 & \textbf{9.89/10.49} & 33.63/34.89 & \textbf{6.47/8.00} & 24.34/29.05 &356.5/208.8\\ 
	\hline
	\multirow{2}{*}{Transformer \cite{attention}} & baseline & 19.13/20.3 & 65.59/67.21 & 10.25/10.72 & 39.3/40.5 & 7.99/9.68 & 26.91/31.26 &330.6/205.6 \\
	 & LANDMARK & \textbf{22.19/23.87} & 65.48/66.92 & \textbf{11.94/12.84} & 36.62/37.73 & \textbf{8.23/10.43} &  \textbf{28.76/32.70} &348.5/207.8 \\ 
	\hline
	\multirow{2}{*}{G-RCNN \cite{grcnn}} & baseline & 16.46/17.28 & 66.27/67.9 & 9.67/10.17 & 39.3/40.5 & 4.89/5.95 & 30.37/34.47 &366.1/207.1\\
	 & LANDMARK & \textbf{19.24/20.51} & 65.76/67.50 & \textbf{11.14/11.62} & \textbf{42.80/43.55} & \textbf{5.52/7.21} & 24.61/29.69 &386.4/209.6   \\ 
	\hline
	\multirow{2}{*}{Motifs \cite{motif}} & baseline & 18.79/19.69 & 66.17/67.72 & 9.39/9.97 & 41.57/42.7 & 6.24/7.54 & 29.82/33.65 &367.1/211.5\\
	 & LANDMARK & \textbf{22.37/23.79} & 59.34/61.16 & \textbf{14.36/15.25} & 33.57/35.02 & \textbf{7.87/10.56} & 24.65/28.87 &389.8/214.4 \\ 
	\hline
	\multirow{2}{*}{BGNN \cite{BGNN}} & baseline & 17.26/18.29 & 66.14/67.65 & 10.30/10.83 & 39.94/41.17 & 6.46/8.22 & 30.90/35.36 &341.9/205.0\\
	 & LANDMARK & \textbf{20.35/22.63} & 55.10/57.16 & \textbf{11.64/12.79} & 33.3/34.89 & \textbf{7.34/8.97} & 24.34/29.05 &360.1/207.2 \\
	 \hline
	\end{tabular}
	}
	\end{table*}

  \subsection{Compatibility  with  unbiased  methods}
  \noindent  The most tricky problem of unbiased SGG strategies is that most of them has demanding application case. Hence, we test our framework's compatibility with other unbiased  methods by  stacking two methods together, the  mRecall \& Recall are  listed in   Table \ref{tab:capability}.  The listed strategies belong to different type, e.g., Data re-sampling: BLS~\cite{BGNN}, logit manipulation: e.g. TDE~\cite{unbias}, and  feature refinement: LANDMARK (ours). According to this table, there are several findings:
  \begin{itemize}
  \item Applying LANDMARK with  other methods is worked. For instance,  BLS  with LANDMARK on BGNN get new SOTA performance. The reason has two: 1) LANDMARK has distinctive semantic feature enhancement strategy, which do not conflict with other methods. 2) Most of unbiased methods are designed to obtain priors from biased prerequisites (e.g., baseline, long-tailed dataset), whereas LANDMARK is baseline-indepandent, that is, baseline  do not affect LANDMARK's inference.  
  \item Only a tiny improvement, or even decrease in mR@K occurred when using BLS and TDE together. For example, Motif+BLS+TDE results in obvious decrease in mR@k. This  suggests that these methods are sensitive to changes of  external circumstances (e.g., sampling distribution, baseline model capability). 
  \item Our network do not sacrifice  Recall a lot. 
  For example, BLS+LANDMARK on Motifs has higher Recall than without LANDMARK.
   However, using BLS+TDE remarkably impair the Recall performance. We speculate that existing unbiased methods do not explore real discrepancies between predicates, but only on increasing likelihood of tail predicates.
  \end{itemize}

  \begin{table*}[t]
    \centering
    \caption{Compatibility test of unbiased methods and LANDMARK  in \%. * means this Baseline model and unbiased method are proposed in the same paper. All  methods are reimplemented on our codebase.}
    \label{tab:capability}
    \resizebox{\linewidth}{!}{%
    \begin{tabular}{ccccccc} 
    \hline
    \multirow{2}{*}{Method} & \multicolumn{2}{c}{PredCls} & \multicolumn{2}{c}{SGCls} & \multicolumn{2}{c}{SGDet}  \\
     & mR@50/100 & R@50/100 & mR@50/100 & R@50/100 & mR@50/100 & R@50/100  \\ 
    \hline
    $\text{Motifs}_{\text{TDE}}$ & 22.87/25.90 & 35.28/41.11 & 13.76/15.36 & 23.88/27.42 & 7.87/9.55 & 9.83/13.06 \\
    $\text{Motifs}_{\text{BLS}}$ & 30.64/32.76 & 54.56/56.27 & 20.21/21.09 & 34.11/35.00 & 13.20/15.97 & 25.27/28.88 \\
    $\text{Motifs}_{\text{BLS+TDE}}$ & 26.19/30.84 & 17.01/18.89  & 17.31/19.28 & 19.63/21.43  & 9.18/11.89 & 8.34/10.27 \\
    $\text{Motifs}_{\text{TDE+LANDMARK}}$ & 24.59/29.27 & 39.24/45.25 & 14.43/17.03 & 28.70/32.28 & 8.98/11.35 & 9.92/12.85 \\
    $\text{Motifs}_{\text{BLS+LANDMARK}}$ & \textbf{33.62/35.69} & \textbf{55.29/56.80} & \textbf{21.43/22.05} & \textbf{36.02/36.85} & \textbf{13.86/16.31} & 25.10/\textbf{29.12} \\ 
    \hline
     $\text{BGNN}_{\text{TDE}}$ &  19.23/21.59   &  32.75/36.45   &  13.09/14.18   &  28.66/31.20 &   8.65/11.01 &   20.36/26.00  \\
    $\text{BGNN}_{\text{BLS}}^{* }$ & 30.64/32.76 & 55.39/56.8 & 16.69/17.83 & 35.70/36.94 & 13.17/15.56 & 23.54/27.56 \\
    $\text{BGNN}_{\text{BLS+TDE}}$ &  31.88/34.89  &  30.37/33.33  &  16.97/18.70  &  21.62/23.52  &  11.60/14.32  &  19.45/24.11  \\
    $\text{BGNN}_{\text{TDE+LANDMARK}}$ &  20.32/22.78   &  50.21/53.85  & 14.56/15.58   &  35.31\textbf{/37.38}  &  9.68/12.41  & \textbf{25.72}/27.98  \\
    $\text{BGNN}_{\text{BLS+LANDMARK}}$ & \textbf{34.41/36.43} & 53.99/55.42 & \textbf{18.27/19.12} & 34.55/35.80 & \textbf{14.08/18.34} & 24.87\textbf{/29.41} \\
    \hline
    \end{tabular}
    }
    \end{table*}


  \begin{figure*}[h]
    \centering
    \includegraphics[width=0.98\textwidth]{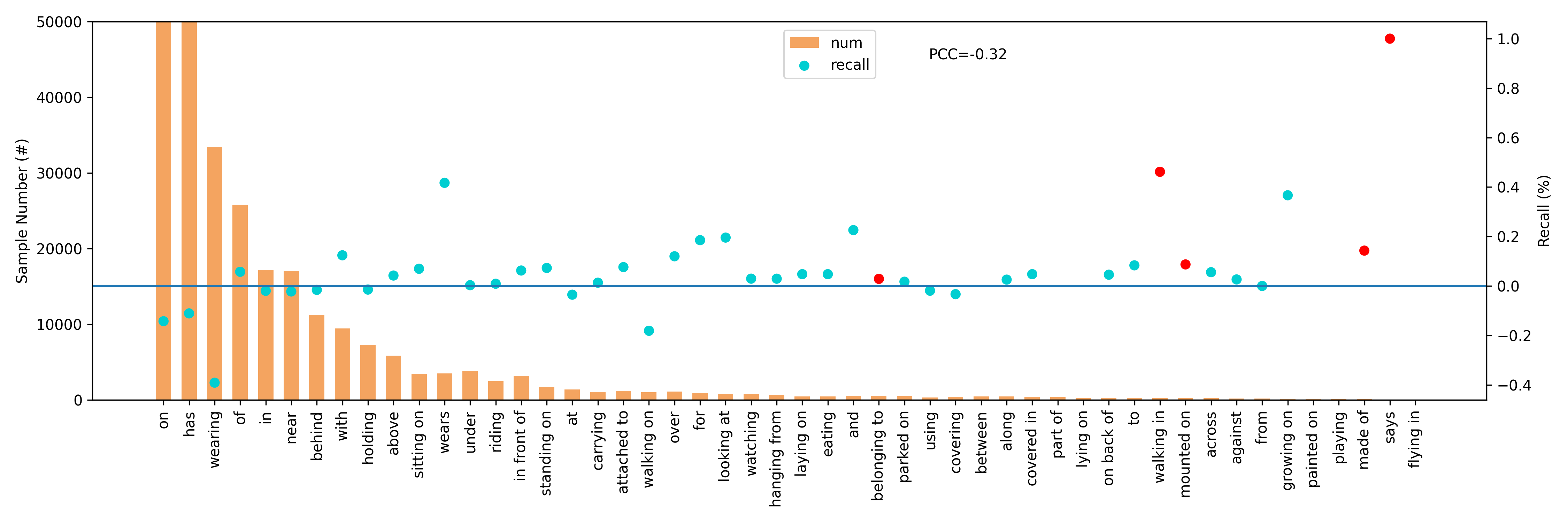}
    \caption{  \textcolor{black}{Data sampling distribution and Recall@100 improvements of LANDMARK on BGNN+BLS over each predicate (on Predcls task).  The red dots indicate BGNN+BLS failed to. PCC is abbreviation of Pearson Correlation Coefficient}}
    \label{predicate}
    \end{figure*}

  \subsection{Analyze Each Predicate:}
  \noindent Shown in Fig. \ref{predicate}, we present VG dataset's class distribution and Recall@100 improvements on Predcls task (BGNN+BLS w/ or w/o ours). Over all 50 predicates, only 11 predicates predicted by baseline are superior to LANDMARK. Besides, there are 6 hard-to-distinguish predicates  (i.e.bitam,\textit{belong to, walking on, mounted on, made of, says})  are recalled by LANDMARK.

  \textcolor{black}{For exploring correlation of Recall improvements  and data distribution,   Pearson Correlation Coefficient (PCC) is used. PCC=0.32 shows weak negative correlation between dataset bias and LANDMARK improvements. Needed to be noticed that some unbiased methods   oversampling tail classes, so that reveal strong negative correlation (e.g., TDE: PCC=-0.56).  Besides, PCC between data  distribution and LANDMARK or BLS on BGNN  (-0.1 vs -0.3) proves both improvements and predictions are robustness to long-tail data.}

  \subsection{Ablation Studies}
  
  \noindent We investigate each LANDMARK component by incrementally adding EEM, LAM, LCM to the
   BGNN+BLS. The results  in Table \ref{tab:ab1} indicate that: 1) each component are helpful for the whole framework, and no confliction between them. It proves three modules extract distinctive semantics from same label inputs.  2) EEM module mainly improves  Predcls than other tasks, which might caused by less accurate position and object label predictions. 3) LCM and LAM consistently promote performances of each task. Because priors from  language context and correlation of word-visual patterns are relatively robust to misclassified but closed semantic object labels.
   \begin{table*}[h]
    \centering
    \caption{Ablation studies on framework structure in three tasks}
    \label{tab:ab1}
    \begin{tabular}{cllcccccc} 
    \hline
    \multicolumn{3}{c}{Methods  } & \multicolumn{1}{c}{PredCls} & \multicolumn{1}{c}{SGCls} & \multicolumn{1}{c}{SGDet} \\ \hline
    \multicolumn{1}{c}{EEM} & \multicolumn{1}{c}{LAM} & \multicolumn{1}{c}{LCM} & mR@50/100 & mR@50/100 & mR@50/100 \\ 
    \hline
     &  &  & 30.64/32.76 & 16.69/17.83 & 13.17/15.56 \\
     \multicolumn{1}{c}{\checkmark} &  &  & 31.75/33.00 & 17.27/18.18 & 13.54/15.92 \\
     \multicolumn{1}{c}{\checkmark} & \multicolumn{1}{c}{\checkmark} &  & 32.06/33.36 & 17.75/18.89 & 14.34/16.42\\
     & \multicolumn{1}{c}{\checkmark} & \multicolumn{1}{c}{\checkmark} & 34.10/36.07 & 18.07/19.01 & 13.89/17.00 \\	
    \multicolumn{1}{c}{\checkmark} & \multicolumn{1}{c}{\checkmark} & \multicolumn{1}{c}{\checkmark} & \textbf{34.41}/\textbf{36.43} & \textbf{18.27}/\textbf{19.12} & \textbf{14.08}/\textbf{18.34} \\
    \hline
    \end{tabular}
    \end{table*}
  \subsection{Analyzation of Experience Estimation module:}
  \noindent Mentioned before, Language Attention Module independently outputs predicate predictions, which is alike Frequency Baseline (FREQ) \cite{motif}. Therefore, we evaluate EEM and FREQ trained on BGNN+BLS+LANDMARK  on evaluation dataset split, results are in Table \ref{tab:eem}.  Except Top-5 on Predcls task, all performance of EEM are outperformed FREQ, and the gap enlarged along with task difficulty increased.  It is attributed to supervision from marginal possibility that alleviates the deficiency of rare subject-object sample.  Besides, position information introduced in EEM  make it  accurate when  inferencing same object pair.
  \begin{table}
    \centering
    \caption{EEM module and FREQ model comparison on baseline Motifs.}
    \label{tab:eem}
    \begin{tabular}{cccc} 
    \hline
    \multirow{2}{*}{Method} & Predcls              & SGCls                         & SGDet                          \\ 
    \cline{2-4}
                            & Top 1/5 R@100        & Top 1/5 R@100                 & Top 1/5 R@100                  \\ 
    \hline
    FREQ \cite{motif}                   & 14.72/\textbf{42.47} & 9.71/27.56                    & 7.66/18.32                     \\
    EEM                     & \textbf{22.74}/40.38 & \textbf{14.76}/\textbf{29.14} & \textbf{14.13}/\textbf{23.43}  \\
    \hline
    \end{tabular}
    \end{table}
 
  \subsection{Analyzation of Language Attention Module:}
  
\begin{figure*}[h]
    \includegraphics[width=0.98\textwidth]{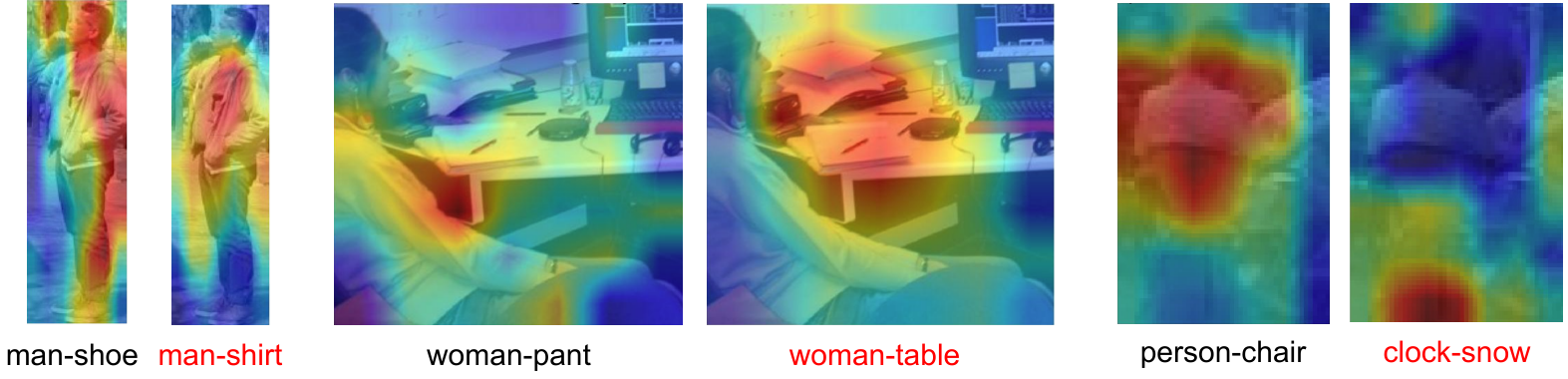}
    
      \caption{Evaluation of LAM. We visualize union area  by given groundtruth subject-object pair and random generalized pair (red words), which shows connection between word pair and particular visual features. }
      
    \label{fig:heatmap}
    \end{figure*}

  \noindent To validate LAM module's effectiveness, we  visualize heatmap of relation representation  $\hat{e}_{ij}$ (Eq. \ref{edge})  trained on BGNN+BLS+LANDMARK  with true or random generated subject-object pair (with red words) in Figure \ref{fig:heatmap}.   Intuitively, we can notice that attention area is correlated with  given subject and object. For instance, in the leftmost two images, attention area transfers from foot  to middle of man's body when object  changes from \textit{shoe} to \textit{shirt}. While given irrelevant words in rightmost two images, e.g., \textit{clock}-\textit{snow}, module seems to interest in top left and bottom area, which suggest LAM can associate words to  related  visual pattern without given coordinates. 
  \label{lmmvisual}

  \subsection{Evaluation of EEM factor}
\begin{figure}[h]
    \centering
  
    \includegraphics[width=0.5\textwidth]{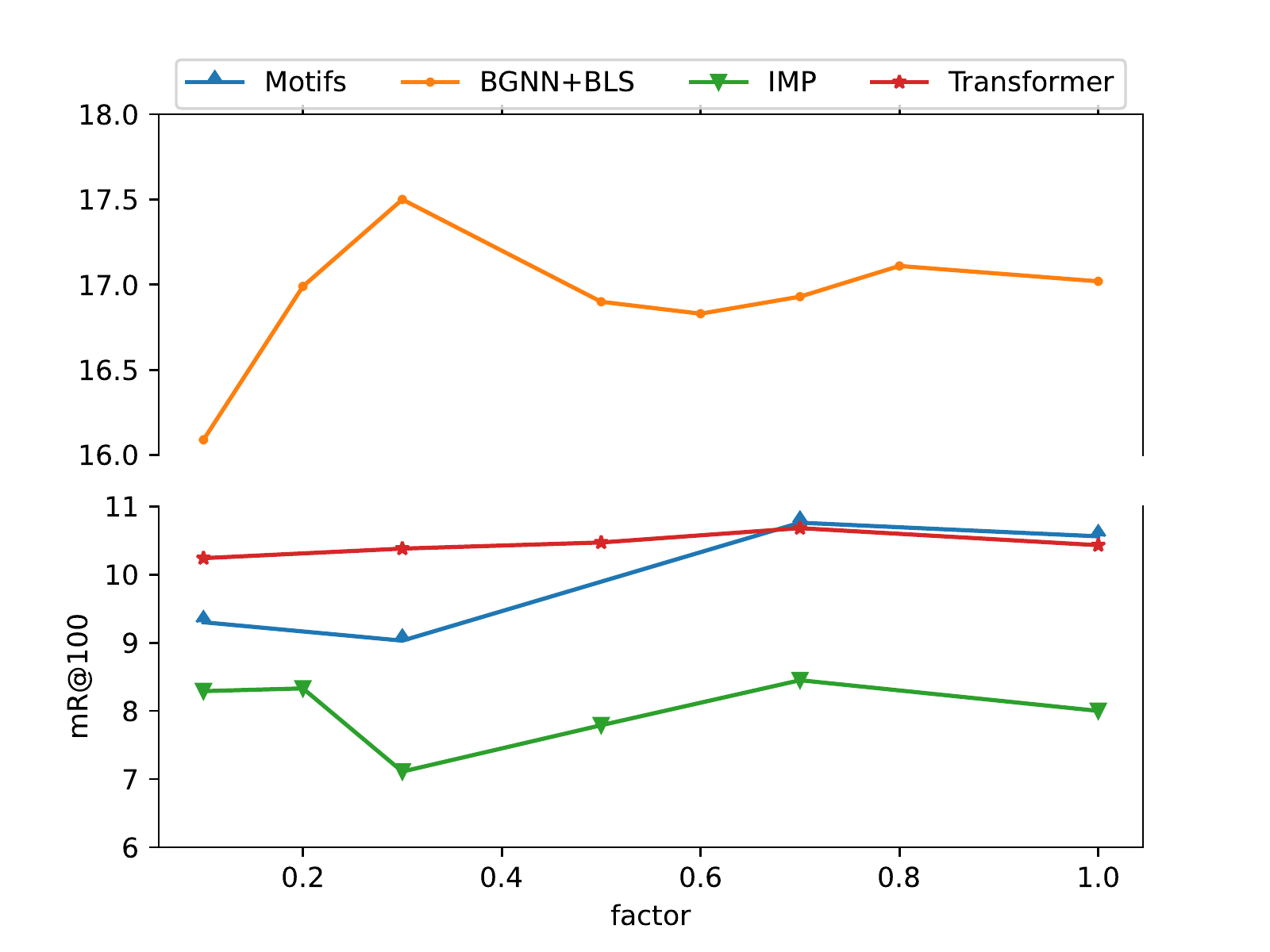}
  
    \caption{ mR@100 performance of SGG models with different EEM's factor on SGDet task.}
    \label{fig:factor}
  \end{figure}

  \noindent   Fig.~\ref{fig:factor} shows  4 model's results of  mRecall@100 on  SGDet task with different EEM's factor $\mu $ in Eq.~\ref{eq:miu}. We find that for  baseline models, factor=0.7 is preferable, for unbiased methods, 0.3 is better. This indicates that calculated marginal  probability mainly records diversified possible predicates, whereas real labels are more likely to be head predicates. 
  
  \begin{figure*}[t]
    \centering
  
    \includegraphics[width=0.98\textwidth]{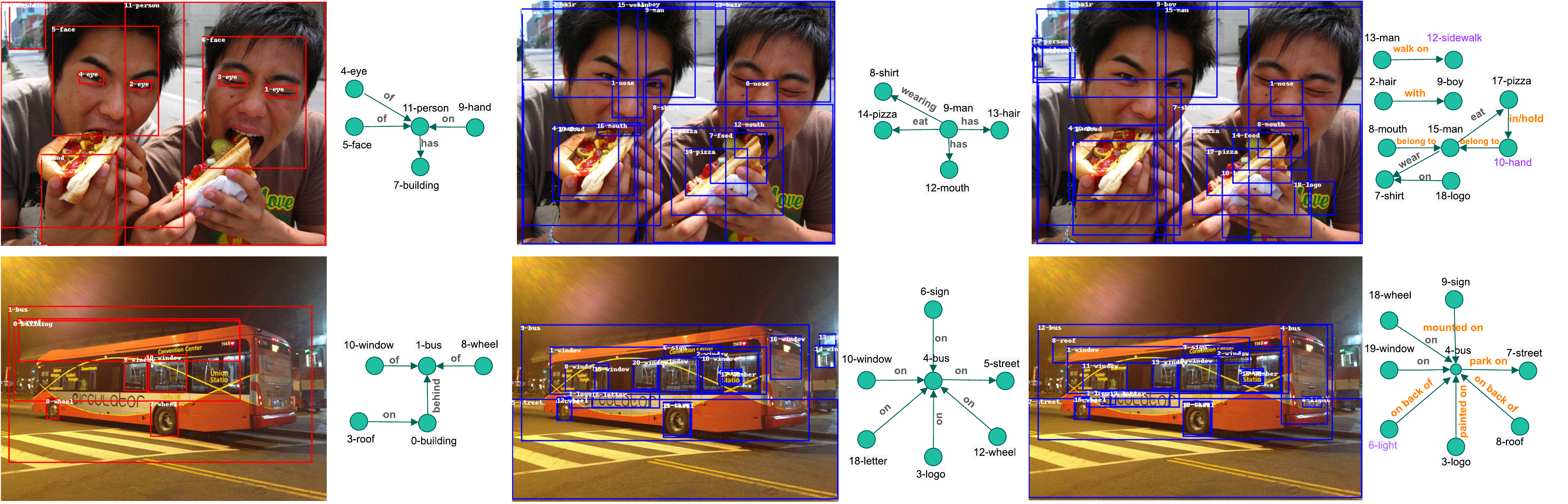}
  
    \caption{Visualization Results: we present scene graphs generated by
      annotations, BGNN with BLS (unbiased model) and LANDMARK in three columns, respectively. The relations
      and entities that neither occur in annotations nor baseline are marked with
      purple and orange.}
    \label{visual}
  \end{figure*}
  \subsection{Qualitative studies}
  \label{qual}
  \noindent We visualize scene graph generation result from annotations, BGNN with BLS (unbiased model), and BGNN+BLS+LANDMARK on Predcls task in Fig.~\ref{visual} 
  Intuitively, annotations and unbiased model tends to predict relationships between ``less informative pairs'' (e.g., \textit{person-eye}, \textit{roof-building} and \textit{light-bus}) that pretty conventional. LANDMARK can further detect relationships between \textit{man-sidewalk} or \textit{roof-bus}. Besides,  LANDMARK focuses on high semantic and positional relationships. For instance,  ``hand hold pizza'' and ``light-on back of-bus'', which prove that our framework successfully learns position relationship  between objects. 

  \section{Conclusion}
  \label{conclusion}
  In this paper, we first point out inadequate language modality utilization in precious SGG methods. Therefore, motivated by language's polysemy, we purpose a  representation enhancement framework (LANDMARK) for SGG task, featured by  multi semantic extraction from object labels. This plug-in network  explores  word-vision correlated pattern and language context from word embedding, and learns predicate distribution from subject-object pair with position. Compared with other unbiased methods, our framework is a new approach from representation refinement perspective. Experiment and analyzes shows constant improvement on Baseline models and compatibility on other unbiased methods. 
\section*{Declarations}

  \subsection*{Competing interests:} 
   No potential conflict of interest was reported by the authors
   \subsection* {Availability of data and materials:} 
  The data that support the findings of this study are available on request from the first author. The results can be implemented  by our GitHub repo.
  \subsection* {Code availability: }
  The code  are openly available in [PySGG-cxg] at [https://github.com/rafa-cxg/PySGG-cxg].

\bibliography{sn-article}


\begin{thebibliography}{41}
\ifx \bisbn   \undefined \def \bisbn  #1{ISBN #1}\fi
\ifx \binits  \undefined \def \binits#1{#1}\fi
\ifx \bauthor  \undefined \def \bauthor#1{#1}\fi
\ifx \batitle  \undefined \def \batitle#1{#1}\fi
\ifx \bjtitle  \undefined \def \bjtitle#1{#1}\fi
\ifx \bvolume  \undefined \def \bvolume#1{\textbf{#1}}\fi
\ifx \byear  \undefined \def \byear#1{#1}\fi
\ifx \bissue  \undefined \def \bissue#1{#1}\fi
\ifx \bfpage  \undefined \def \bfpage#1{#1}\fi
\ifx \blpage  \undefined \def \blpage #1{#1}\fi
\ifx \burl  \undefined \def \burl#1{\textsf{#1}}\fi
\ifx \doiurl  \undefined \def \doiurl#1{\url{https://doi.org/#1}}\fi
\ifx \betal  \undefined \def \betal{\textit{et al.}}\fi
\ifx \binstitute  \undefined \def \binstitute#1{#1}\fi
\ifx \binstitutionaled  \undefined \def \binstitutionaled#1{#1}\fi
\ifx \bctitle  \undefined \def \bctitle#1{#1}\fi
\ifx \beditor  \undefined \def \beditor#1{#1}\fi
\ifx \bpublisher  \undefined \def \bpublisher#1{#1}\fi
\ifx \bbtitle  \undefined \def \bbtitle#1{#1}\fi
\ifx \bedition  \undefined \def \bedition#1{#1}\fi
\ifx \bseriesno  \undefined \def \bseriesno#1{#1}\fi
\ifx \blocation  \undefined \def \blocation#1{#1}\fi
\ifx \bsertitle  \undefined \def \bsertitle#1{#1}\fi
\ifx \bsnm \undefined \def \bsnm#1{#1}\fi
\ifx \bsuffix \undefined \def \bsuffix#1{#1}\fi
\ifx \bparticle \undefined \def \bparticle#1{#1}\fi
\ifx \barticle \undefined \def \barticle#1{#1}\fi
\bibcommenthead
\ifx \bconfdate \undefined \def \bconfdate #1{#1}\fi
\ifx \botherref \undefined \def \botherref #1{#1}\fi
\ifx \url \undefined \def \url#1{\textsf{#1}}\fi
\ifx \bchapter \undefined \def \bchapter#1{#1}\fi
\ifx \bbook \undefined \def \bbook#1{#1}\fi
\ifx \bcomment \undefined \def \bcomment#1{#1}\fi
\ifx \oauthor \undefined \def \oauthor#1{#1}\fi
\ifx \citeauthoryear \undefined \def \citeauthoryear#1{#1}\fi
\ifx \endbibitem  \undefined \def \endbibitem {}\fi
\ifx \bconflocation  \undefined \def \bconflocation#1{#1}\fi
\ifx \arxivurl  \undefined \def \arxivurl#1{\textsf{#1}}\fi
\csname PreBibitemsHook\endcsname

\bibitem{ic1}
\begin{bchapter}
\bauthor{\bsnm{Gu}, \binits{J.}},
\bauthor{\bsnm{Joty}, \binits{S.}},
\bauthor{\bsnm{Cai}, \binits{J.}},
\bauthor{\bsnm{Zhao}, \binits{H.}},
\bauthor{\bsnm{Yang}, \binits{X.}},
\bauthor{\bsnm{Wang}, \binits{G.}}:
\bctitle{Unpaired image captioning via scene graph alignments}.
In: \bbtitle{Proceedings of the IEEE/CVF International Conference on Computer
  Vision},
pp. \bfpage{10323}--\blpage{10332}
(\byear{2019})
\end{bchapter}
\endbibitem

\bibitem{ic3}
\begin{barticle}
\bauthor{\bsnm{Xu}, \binits{N.}},
\bauthor{\bsnm{Liu}, \binits{A.-A.}},
\bauthor{\bsnm{Liu}, \binits{J.}},
\bauthor{\bsnm{Nie}, \binits{W.}},
\bauthor{\bsnm{Su}, \binits{Y.}}:
\batitle{Scene graph captioner: Image captioning based on structural visual
  representation}.
\bjtitle{Journal of Visual Communication and Image Representation}
\bvolume{58},
\bfpage{477}--\blpage{485}
(\byear{2019})
\end{barticle}
\endbibitem

\bibitem{vqa3}
\begin{botherref}
\oauthor{\bsnm{Zhang}, \binits{C.}},
\oauthor{\bsnm{Chao}, \binits{W.-L.}},
\oauthor{\bsnm{Xuan}, \binits{D.}}:
An empirical study on leveraging scene graphs for visual question answering.
arXiv preprint arXiv:1907.12133
(2019)
\end{botherref}
\endbibitem

\bibitem{vqa2}
\begin{bchapter}
\bauthor{\bsnm{Shi}, \binits{J.}},
\bauthor{\bsnm{Zhang}, \binits{H.}},
\bauthor{\bsnm{Li}, \binits{J.}}:
\bctitle{Explainable and explicit visual reasoning over scene graphs}.
In: \bbtitle{Proceedings of the IEEE/CVF Conference on Computer Vision and
  Pattern Recognition},
pp. \bfpage{8376}--\blpage{8384}
(\byear{2019})
\end{bchapter}
\endbibitem

\bibitem{vqa1}
\begin{bchapter}
\bauthor{\bsnm{Teney}, \binits{D.}},
\bauthor{\bsnm{Liu}, \binits{L.}},
\bauthor{\bparticle{van} \bsnm{Den~Hengel}, \binits{A.}}:
\bctitle{Graph-structured representations for visual question answering}.
In: \bbtitle{Proceedings of the IEEE Conference on Computer Vision and Pattern
  Recognition},
pp. \bfpage{1}--\blpage{9}
(\byear{2017})
\end{bchapter}
\endbibitem

\bibitem{va1}
\begin{bchapter}
\bauthor{\bsnm{Teng}, \binits{Y.}},
\bauthor{\bsnm{Wang}, \binits{L.}},
\bauthor{\bsnm{Li}, \binits{Z.}},
\bauthor{\bsnm{Wu}, \binits{G.}}:
\bctitle{Target adaptive context aggregation for video scene graph generation}.
In: \bbtitle{Proceedings of the IEEE/CVF International Conference on Computer
  Vision},
pp. \bfpage{13688}--\blpage{13697}
(\byear{2021})
\end{bchapter}
\endbibitem

\bibitem{BGNN}
\begin{bchapter}
\bauthor{\bsnm{Li}, \binits{R.}},
\bauthor{\bsnm{Zhang}, \binits{S.}},
\bauthor{\bsnm{Wan}, \binits{B.}},
\bauthor{\bsnm{He}, \binits{X.}}:
\bctitle{Bipartite graph network with adaptive message passing for unbiased
  scene graph generation}.
In: \bbtitle{Proceedings of the IEEE/CVF Conference on Computer Vision and
  Pattern Recognition (CVPR)},
pp. \bfpage{11109}--\blpage{11119}
(\byear{2021})
\end{bchapter}
\endbibitem

\bibitem{stracked}
\begin{botherref}
\oauthor{\bsnm{Dong}, \binits{X.}},
\oauthor{\bsnm{Gan}, \binits{T.}},
\oauthor{\bsnm{Song}, \binits{X.}},
\oauthor{\bsnm{Wu}, \binits{J.}},
\oauthor{\bsnm{Cheng}, \binits{Y.}},
\oauthor{\bsnm{Nie}, \binits{L.}}:
Stacked hybrid-attention and group collaborative learning for unbiased scene
  graph generation.
arXiv preprint arXiv:2203.09811
(2022)
\end{botherref}
\endbibitem

\bibitem{cogtree}
\begin{botherref}
\oauthor{\bsnm{Yu}, \binits{J.}},
\oauthor{\bsnm{Chai}, \binits{Y.}},
\oauthor{\bsnm{Wang}, \binits{Y.}},
\oauthor{\bsnm{Hu}, \binits{Y.}},
\oauthor{\bsnm{Wu}, \binits{Q.}}:
Cogtree: Cognition tree loss for unbiased scene graph generation.
arXiv preprint arXiv:2009.07526
(2020)
\end{botherref}
\endbibitem

\bibitem{finegrain}
\begin{bchapter}
\bauthor{\bsnm{Lyu}, \binits{X.}},
\bauthor{\bsnm{Gao}, \binits{L.}},
\bauthor{\bsnm{Guo}, \binits{Y.}},
\bauthor{\bsnm{Zhao}, \binits{Z.}},
\bauthor{\bsnm{Huang}, \binits{H.}},
\bauthor{\bsnm{Shen}, \binits{H.T.}},
\bauthor{\bsnm{Song}, \binits{J.}}:
\bctitle{Fine-grained predicates learning for scene graph generation}.
In: \bbtitle{Proceedings of the IEEE/CVF Conference on Computer Vision and
  Pattern Recognition},
pp. \bfpage{19467}--\blpage{19475}
(\byear{2022})
\end{bchapter}
\endbibitem

\bibitem{unbias}
\begin{bchapter}
\bauthor{\bsnm{Tang}, \binits{K.}},
\bauthor{\bsnm{Niu}, \binits{Y.}},
\bauthor{\bsnm{Huang}, \binits{J.}},
\bauthor{\bsnm{Shi}, \binits{J.}},
\bauthor{\bsnm{Zhang}, \binits{H.}}:
\bctitle{Unbiased scene graph generation from biased training}.
In: \bbtitle{Proceedings of the IEEE/CVF Conference on Computer Vision and
  Pattern Recognition},
pp. \bfpage{3716}--\blpage{3725}
(\byear{2020})
\end{bchapter}
\endbibitem

\bibitem{resistance}
\begin{botherref}
\oauthor{\bsnm{Chen}, \binits{C.}},
\oauthor{\bsnm{Zhan}, \binits{Y.}},
\oauthor{\bsnm{Yu}, \binits{B.}},
\oauthor{\bsnm{Liu}, \binits{L.}},
\oauthor{\bsnm{Luo}, \binits{Y.}},
\oauthor{\bsnm{Du}, \binits{B.}}:
Resistance training using prior bias: toward unbiased scene graph generation.
arXiv preprint arXiv:2201.06794
(2022)
\end{botherref}
\endbibitem

\bibitem{motif}
\begin{bchapter}
\bauthor{\bsnm{Zellers}, \binits{R.}},
\bauthor{\bsnm{Yatskar}, \binits{M.}},
\bauthor{\bsnm{Thomson}, \binits{S.}},
\bauthor{\bsnm{Choi}, \binits{Y.}}:
\bctitle{Neural motifs: Scene graph parsing with global context}.
In: \bbtitle{Proceedings of the IEEE Conference on Computer Vision and Pattern
  Recognition},
pp. \bfpage{5831}--\blpage{5840}
(\byear{2018})
\end{bchapter}
\endbibitem

\bibitem{attentiontranslation}
\begin{bchapter}
\bauthor{\bsnm{Gkanatsios}, \binits{N.}},
\bauthor{\bsnm{Pitsikalis}, \binits{V.}},
\bauthor{\bsnm{Koutras}, \binits{P.}},
\bauthor{\bsnm{Maragos}, \binits{P.}}:
\bctitle{Attention-translation-relation network for scalable scene graph
  generation}.
In: \bbtitle{Proceedings of the IEEE/CVF International Conference on Computer
  Vision Workshops},
pp. \bfpage{0}--\blpage{0}
(\byear{2019})
\end{bchapter}
\endbibitem

\bibitem{sgnls}
\begin{bchapter}
\bauthor{\bsnm{Zhong}, \binits{Y.}},
\bauthor{\bsnm{Shi}, \binits{J.}},
\bauthor{\bsnm{Yang}, \binits{J.}},
\bauthor{\bsnm{Xu}, \binits{C.}},
\bauthor{\bsnm{Li}, \binits{Y.}}:
\bctitle{Learning to generate scene graph from natural language supervision}.
In: \bbtitle{Proceedings of the IEEE/CVF International Conference on Computer
  Vision},
pp. \bfpage{1823}--\blpage{1834}
(\byear{2021})
\end{bchapter}
\endbibitem

\bibitem{gcn}
\begin{botherref}
\oauthor{\bsnm{Kipf}, \binits{T.N.}},
\oauthor{\bsnm{Welling}, \binits{M.}}:
Semi-supervised classification with graph convolutional networks.
arXiv preprint arXiv:1609.02907
(2016)
\end{botherref}
\endbibitem

\bibitem{Gps-net}
\begin{bchapter}
\bauthor{\bsnm{Lin}, \binits{X.}},
\bauthor{\bsnm{Ding}, \binits{C.}},
\bauthor{\bsnm{Zeng}, \binits{J.}},
\bauthor{\bsnm{Tao}, \binits{D.}}:
\bctitle{Gps-net: Graph property sensing network for scene graph generation}.
In: \bbtitle{Proceedings of the IEEE/CVF Conference on Computer Vision and
  Pattern Recognition},
pp. \bfpage{3746}--\blpage{3753}
(\byear{2020})
\end{bchapter}
\endbibitem

\bibitem{mem}
\begin{bchapter}
\bauthor{\bsnm{Wang}, \binits{W.}},
\bauthor{\bsnm{Wang}, \binits{R.}},
\bauthor{\bsnm{Shan}, \binits{S.}},
\bauthor{\bsnm{Chen}, \binits{X.}}:
\bctitle{Exploring context and visual pattern of relationship for scene graph
  generation}.
In: \bbtitle{Proceedings of the IEEE/CVF Conference on Computer Vision and
  Pattern Recognition (CVPR)}
(\byear{2019})
\end{bchapter}
\endbibitem

\bibitem{linknet}
\begin{botherref}
\oauthor{\bsnm{Woo}, \binits{S.}},
\oauthor{\bsnm{Kim}, \binits{D.}},
\oauthor{\bsnm{Cho}, \binits{D.}},
\oauthor{\bsnm{Kweon}, \binits{I.S.}}:
Linknet: Relational embedding for scene graph.
Advances in Neural Information Processing Systems
\textbf{31}
(2018)
\end{botherref}
\endbibitem

\bibitem{factorizable}
\begin{bchapter}
\bauthor{\bsnm{Li}, \binits{Y.}},
\bauthor{\bsnm{Ouyang}, \binits{W.}},
\bauthor{\bsnm{Zhou}, \binits{B.}},
\bauthor{\bsnm{Shi}, \binits{J.}},
\bauthor{\bsnm{Zhang}, \binits{C.}},
\bauthor{\bsnm{Wang}, \binits{X.}}:
\bctitle{Factorizable net: an efficient subgraph-based framework for scene
  graph generation}.
In: \bbtitle{Proceedings of the European Conference on Computer Vision (ECCV)},
pp. \bfpage{335}--\blpage{351}
(\byear{2018})
\end{bchapter}
\endbibitem

\bibitem{structure}
\begin{bchapter}
\bauthor{\bsnm{Liu}, \binits{Y.}},
\bauthor{\bsnm{Wang}, \binits{R.}},
\bauthor{\bsnm{Shan}, \binits{S.}},
\bauthor{\bsnm{Chen}, \binits{X.}}:
\bctitle{Structure inference net: Object detection using scene-level context
  and instance-level relationships}.
In: \bbtitle{Proceedings of the IEEE Conference on Computer Vision and Pattern
  Recognition},
pp. \bfpage{6985}--\blpage{6994}
(\byear{2018})
\end{bchapter}
\endbibitem

\bibitem{phrases}
\begin{bchapter}
\bauthor{\bsnm{Li}, \binits{Y.}},
\bauthor{\bsnm{Ouyang}, \binits{W.}},
\bauthor{\bsnm{Zhou}, \binits{B.}},
\bauthor{\bsnm{Wang}, \binits{K.}},
\bauthor{\bsnm{Wang}, \binits{X.}}:
\bctitle{Scene graph generation from objects, phrases and region captions}.
In: \bbtitle{Proceedings of the IEEE International Conference on Computer
  Vision},
pp. \bfpage{1261}--\blpage{1270}
(\byear{2017})
\end{bchapter}
\endbibitem

\bibitem{lstm}
\begin{barticle}
\bauthor{\bsnm{Hochreiter}, \binits{S.}},
\bauthor{\bsnm{Schmidhuber}, \binits{J.}}:
\batitle{Long short-term memory}.
\bjtitle{Neural computation}
\bvolume{9}(\bissue{8}),
\bfpage{1735}--\blpage{1780}
(\byear{1997})
\end{barticle}
\endbibitem

\bibitem{gru}
\begin{botherref}
\oauthor{\bsnm{Cho}, \binits{K.}},
\oauthor{\bsnm{Van~Merri{\"e}nboer}, \binits{B.}},
\oauthor{\bsnm{Gulcehre}, \binits{C.}},
\oauthor{\bsnm{Bahdanau}, \binits{D.}},
\oauthor{\bsnm{Bougares}, \binits{F.}},
\oauthor{\bsnm{Schwenk}, \binits{H.}},
\oauthor{\bsnm{Bengio}, \binits{Y.}}:
Learning phrase representations using rnn encoder-decoder for statistical
  machine translation.
arXiv preprint arXiv:1406.1078
(2014)
\end{botherref}
\endbibitem

\bibitem{attention}
\begin{bchapter}
\bauthor{\bsnm{Vaswani}, \binits{A.}},
\bauthor{\bsnm{Shazeer}, \binits{N.}},
\bauthor{\bsnm{Parmar}, \binits{N.}},
\bauthor{\bsnm{Uszkoreit}, \binits{J.}},
\bauthor{\bsnm{Jones}, \binits{L.}},
\bauthor{\bsnm{Gomez}, \binits{A.N.}},
\bauthor{\bsnm{Kaiser}, \binits{{\L}.}},
\bauthor{\bsnm{Polosukhin}, \binits{I.}}:
\bctitle{Attention is all you need}.
In: \bbtitle{Advances in Neural Information Processing Systems},
pp. \bfpage{5998}--\blpage{6008}
(\byear{2017})
\end{bchapter}
\endbibitem

\bibitem{energy}
\begin{bchapter}
\bauthor{\bsnm{Suhail}, \binits{M.}},
\bauthor{\bsnm{Mittal}, \binits{A.}},
\bauthor{\bsnm{Siddiquie}, \binits{B.}},
\bauthor{\bsnm{Broaddus}, \binits{C.}},
\bauthor{\bsnm{Eledath}, \binits{J.}},
\bauthor{\bsnm{Medioni}, \binits{G.}},
\bauthor{\bsnm{Sigal}, \binits{L.}}:
\bctitle{Energy-based learning for scene graph generation}.
In: \bbtitle{Proceedings of the IEEE/CVF Conference on Computer Vision and
  Pattern Recognition},
pp. \bfpage{13936}--\blpage{13945}
(\byear{2021})
\end{bchapter}
\endbibitem

\bibitem{grcnn}
\begin{bchapter}
\bauthor{\bsnm{Yang}, \binits{J.}},
\bauthor{\bsnm{Lu}, \binits{J.}},
\bauthor{\bsnm{Lee}, \binits{S.}},
\bauthor{\bsnm{Batra}, \binits{D.}},
\bauthor{\bsnm{Parikh}, \binits{D.}}:
\bctitle{Graph r-cnn for scene graph generation}.
In: \bbtitle{Proceedings of the European Conference on Computer Vision (ECCV)},
pp. \bfpage{670}--\blpage{685}
(\byear{2018})
\end{bchapter}
\endbibitem

\bibitem{drnet}
\begin{bchapter}
\bauthor{\bsnm{Dai}, \binits{B.}},
\bauthor{\bsnm{Zhang}, \binits{Y.}},
\bauthor{\bsnm{Lin}, \binits{D.}}:
\bctitle{Detecting visual relationships with deep relational networks}.
In: \bbtitle{Proceedings of the IEEE Conference on Computer Vision and Pattern
  Recognition},
pp. \bfpage{3076}--\blpage{3086}
(\byear{2017})
\end{bchapter}
\endbibitem

\bibitem{kern}
\begin{bchapter}
\bauthor{\bsnm{Chen}, \binits{T.}},
\bauthor{\bsnm{Yu}, \binits{W.}},
\bauthor{\bsnm{Chen}, \binits{R.}},
\bauthor{\bsnm{Lin}, \binits{L.}}:
\bctitle{Knowledge-embedded routing network for scene graph generation}.
In: \bbtitle{Proceedings of the IEEE/CVF Conference on Computer Vision and
  Pattern Recognition},
pp. \bfpage{6163}--\blpage{6171}
(\byear{2019})
\end{bchapter}
\endbibitem

\bibitem{vctree}
\begin{bchapter}
\bauthor{\bsnm{Tang}, \binits{K.}},
\bauthor{\bsnm{Zhang}, \binits{H.}},
\bauthor{\bsnm{Wu}, \binits{B.}},
\bauthor{\bsnm{Luo}, \binits{W.}},
\bauthor{\bsnm{Liu}, \binits{W.}}:
\bctitle{Learning to compose dynamic tree structures for visual contexts}.
In: \bbtitle{Proceedings of the IEEE/CVF Conference on Computer Vision and
  Pattern Recognition},
pp. \bfpage{6619}--\blpage{6628}
(\byear{2019})
\end{bchapter}
\endbibitem

\bibitem{lp}
\begin{bchapter}
\bauthor{\bsnm{Lu}, \binits{C.}},
\bauthor{\bsnm{Krishna}, \binits{R.}},
\bauthor{\bsnm{Bernstein}, \binits{M.}},
\bauthor{\bsnm{Fei-Fei}, \binits{L.}}:
\bctitle{Visual relationship detection with language priors}.
In: \bbtitle{European Conference on Computer Vision},
pp. \bfpage{852}--\blpage{869}
(\byear{2016}).
\bcomment{Springer}
\end{bchapter}
\endbibitem

\bibitem{bridge}
\begin{bchapter}
\bauthor{\bsnm{Zareian}, \binits{A.}},
\bauthor{\bsnm{Karaman}, \binits{S.}},
\bauthor{\bsnm{Chang}, \binits{S.-F.}}:
\bctitle{Bridging knowledge graphs to generate scene graphs}.
In: \bbtitle{European Conference on Computer Vision},
pp. \bfpage{606}--\blpage{623}
(\byear{2020}).
\bcomment{Springer}
\end{bchapter}
\endbibitem

\bibitem{commonsense}
\begin{bchapter}
\bauthor{\bsnm{Zareian}, \binits{A.}},
\bauthor{\bsnm{Wang}, \binits{Z.}},
\bauthor{\bsnm{You}, \binits{H.}},
\bauthor{\bsnm{Chang}, \binits{S.-F.}}:
\bctitle{Learning visual commonsense for robust scene graph generation}.
In: \bbtitle{European Conference on Computer Vision},
pp. \bfpage{642}--\blpage{657}
(\byear{2020}).
\bcomment{Springer}
\end{bchapter}
\endbibitem

\bibitem{resnet}
\begin{bchapter}
\bauthor{\bsnm{He}, \binits{K.}},
\bauthor{\bsnm{Zhang}, \binits{X.}},
\bauthor{\bsnm{Ren}, \binits{S.}},
\bauthor{\bsnm{Sun}, \binits{J.}}:
\bctitle{Deep residual learning for image recognition}.
In: \bbtitle{Proceedings of the IEEE Conference on Computer Vision and Pattern
  Recognition},
pp. \bfpage{770}--\blpage{778}
(\byear{2016})
\end{bchapter}
\endbibitem

\bibitem{seq2seq}
\begin{botherref}
\oauthor{\bsnm{Lu}, \binits{Y.}},
\oauthor{\bsnm{Rai}, \binits{H.}},
\oauthor{\bsnm{Chang}, \binits{J.}},
\oauthor{\bsnm{Knyazev}, \binits{B.}},
\oauthor{\bsnm{Yu}, \binits{G.}},
\oauthor{\bsnm{Shekhar}, \binits{S.}},
\oauthor{\bsnm{Taylor}, \binits{G.W.}},
\oauthor{\bsnm{Volkovs}, \binits{M.}}:
Context-aware scene graph generation with seq2seq transformers
\end{botherref}
\endbibitem

\bibitem{visualgenome}
\begin{barticle}
\bauthor{\bsnm{Krishna}, \binits{R.}},
\bauthor{\bsnm{Zhu}, \binits{Y.}},
\bauthor{\bsnm{Groth}, \binits{O.}},
\bauthor{\bsnm{Johnson}, \binits{J.}},
\bauthor{\bsnm{Hata}, \binits{K.}},
\bauthor{\bsnm{Kravitz}, \binits{J.}},
\bauthor{\bsnm{Chen}, \binits{S.}},
\bauthor{\bsnm{Kalantidis}, \binits{Y.}},
\bauthor{\bsnm{Li}, \binits{L.-J.}},
\bauthor{\bsnm{Shamma}, \binits{D.A.}}, \betal:
\batitle{Visual genome: Connecting language and vision using crowdsourced dense
  image annotations}.
\bjtitle{International journal of computer vision}
\bvolume{123}(\bissue{1}),
\bfpage{32}--\blpage{73}
(\byear{2017})
\end{barticle}
\endbibitem

\bibitem{imp}
\begin{bchapter}
\bauthor{\bsnm{Xu}, \binits{D.}},
\bauthor{\bsnm{Zhu}, \binits{Y.}},
\bauthor{\bsnm{Choy}, \binits{C.B.}},
\bauthor{\bsnm{Fei-Fei}, \binits{L.}}:
\bctitle{Scene graph generation by iterative message passing}.
In: \bbtitle{Proceedings of the IEEE Conference on Computer Vision and Pattern
  Recognition},
pp. \bfpage{5410}--\blpage{5419}
(\byear{2017})
\end{bchapter}
\endbibitem

\bibitem{pixels}
\begin{botherref}
\oauthor{\bsnm{Newell}, \binits{A.}},
\oauthor{\bsnm{Deng}, \binits{J.}}:
Pixels to graphs by associative embedding.
Advances in neural information processing systems
\textbf{30}
(2017)
\end{botherref}
\endbibitem

\bibitem{rcnn}
\begin{botherref}
\oauthor{\bsnm{Ren}, \binits{S.}},
\oauthor{\bsnm{He}, \binits{K.}},
\oauthor{\bsnm{Girshick}, \binits{R.}},
\oauthor{\bsnm{Sun}, \binits{J.}}:
Faster r-cnn: Towards real-time object detection with region proposal networks.
Advances in neural information processing systems
\textbf{28}
(2015)
\end{botherref}
\endbibitem

\bibitem{maskrcnn}
\begin{bchapter}
\bauthor{\bsnm{He}, \binits{K.}},
\bauthor{\bsnm{Gkioxari}, \binits{G.}},
\bauthor{\bsnm{Doll{\'a}r}, \binits{P.}},
\bauthor{\bsnm{Girshick}, \binits{R.}}:
\bctitle{Mask r-cnn}.
In: \bbtitle{Proceedings of the IEEE International Conference on Computer
  Vision},
pp. \bfpage{2961}--\blpage{2969}
(\byear{2017})
\end{bchapter}
\endbibitem

\bibitem{glove}
\begin{bchapter}
\bauthor{\bsnm{Pennington}, \binits{J.}},
\bauthor{\bsnm{Socher}, \binits{R.}},
\bauthor{\bsnm{Manning}, \binits{C.D.}}:
\bctitle{Glove: Global vectors for word representation}.
In: \bbtitle{Proceedings of the 2014 Conference on Empirical Methods in Natural
  Language Processing (EMNLP)},
pp. \bfpage{1532}--\blpage{1543}
(\byear{2014})
\end{bchapter}
\endbibitem

\end{thebibliography}


\end{document}